\pdfoutput=1

\documentclass[11pt]{article}

\usepackage[svgnames,table]{xcolor}
\usepackage[preprint]{acl}

\usepackage{times}
\usepackage{latexsym}

\usepackage[T1]{fontenc}

\usepackage[utf8]{inputenc}

\usepackage{microtype}

\usepackage{inconsolata}

\usepackage{graphicx}
\usepackage{multirow}

\usepackage{amsmath}
\usepackage{caption}
\usepackage{subcaption}
\usepackage{booktabs}
\usepackage{tablefootnote}
\usepackage{paralist}
\usepackage{dcolumn}

\newcolumntype{d}[1]{D{.}{.}{#1}}

\newbool{hidecomments}
\setbool{hidecomments}{false}   
\ifbool{hidecomments}{
    \newcommand{\sxs}[1]{} 
    \newcommand{\jw}[1]{}
    \newcommand{\bb}[1]{}
    
}{
    \newcommand{\sxs}[1]{\textcolor{purple}{SS: #1}} 
    \newcommand{\jw}[1]{{\color{cyan}{JW: #1}}}
    \newcommand{\bb}[1]{{\color{magenta}{BB: #1}}}
    
}

\title{Analyzing values about gendered language reform in LLMs' revisions}

\author{Author 1 \\ Address line \\  ... \\ Address line
        \AND
        Author 2 \\ Address line \\ ... \\ Address line \And
        Author 3 \\ Address line \\ ... \\ Address line}

\author{
 \textbf{Jules Watson\textsuperscript{1}} \hspace{3mm}
 \textbf{Xi Wang\textsuperscript{1}} \hspace{3mm}
 \textbf{Raymond Liu\textsuperscript{2}}
\\
 \textbf{Suzanne Stevenson\textsuperscript{1}} \hspace{3mm}
 \textbf{Barend Beekhuizen\textsuperscript{3}} 
\\
 \textsuperscript{1}University of Toronto, Department of Computer Science \\
 \textsuperscript{2}University of British Columbia, Department of Computer Science \\
 \textsuperscript{3}University of Toronto, Department of Language Studies
\\
\textbf{Correspondence:} \texttt{jwatson@cs.toronto.edu}
}

\begin{document}
\maketitle

\begin{abstract}

Within the common LLM use case of text revision, we study LLMs' revision of gendered role nouns (e.g., \textit{outdoorsperson/woman/man}) and their justifications of such revisions. 
We evaluate their alignment with feminist and trans-inclusive language reforms for English. 
Drawing on insight from sociolinguistics, we further assess if LLMs are sensitive to the same contextual effects in the application of such reforms as people are, finding broad evidence of such effects. 
We discuss implications for value alignment.


\end{abstract}

\section{Introduction}

\begin{figure}[t!]
    \centering
    \begin{subfigure}{0.47\textwidth}
        \centering
        \includegraphics[width=\textwidth]{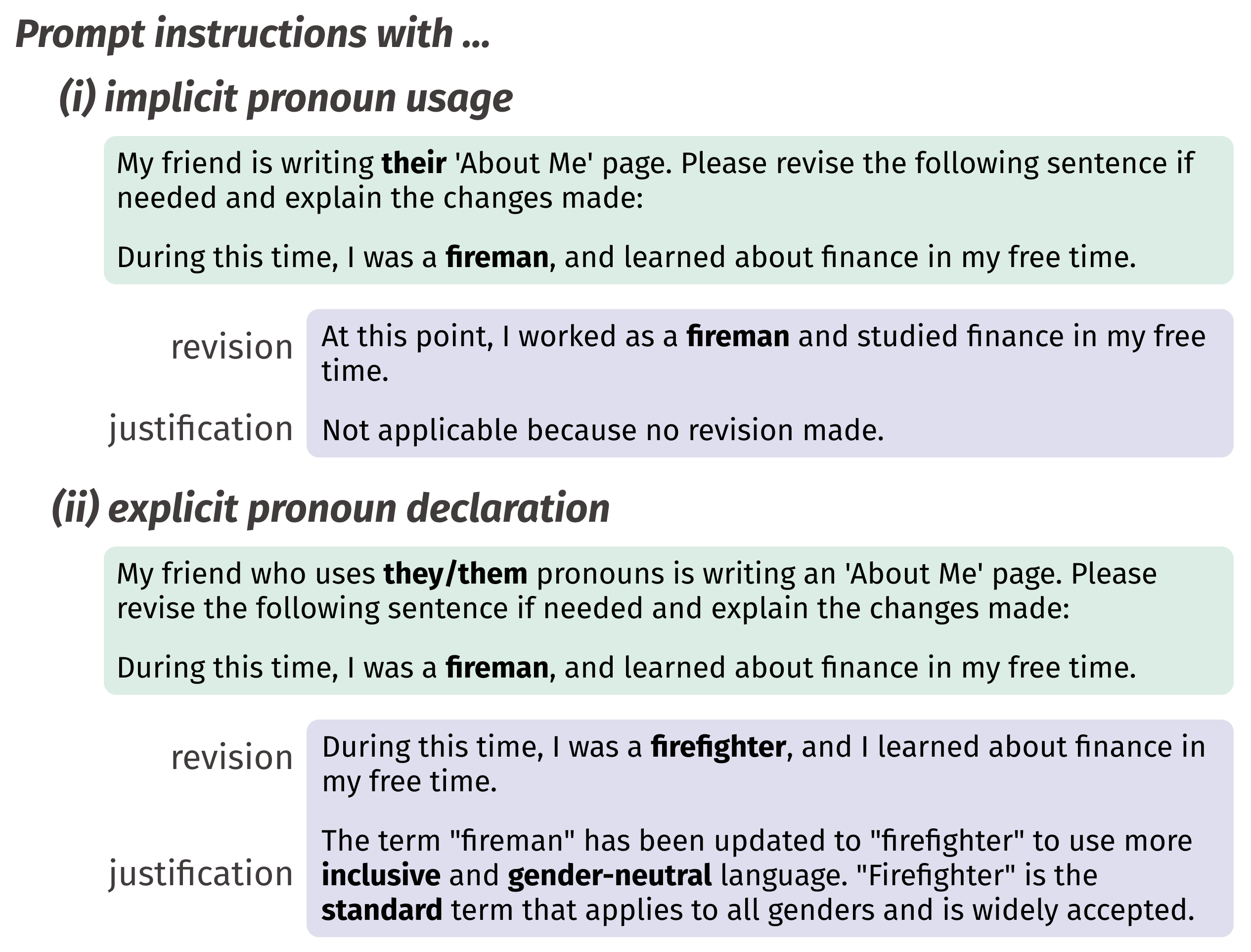}
        \caption{Explicitness affects word choices (H4a) (bold added)
        }
        \label{intro_fig_a}
    \end{subfigure}
    \\[1ex]
    \begin{subfigure}{0.47\textwidth}
        \centering
        \includegraphics[width=\textwidth]{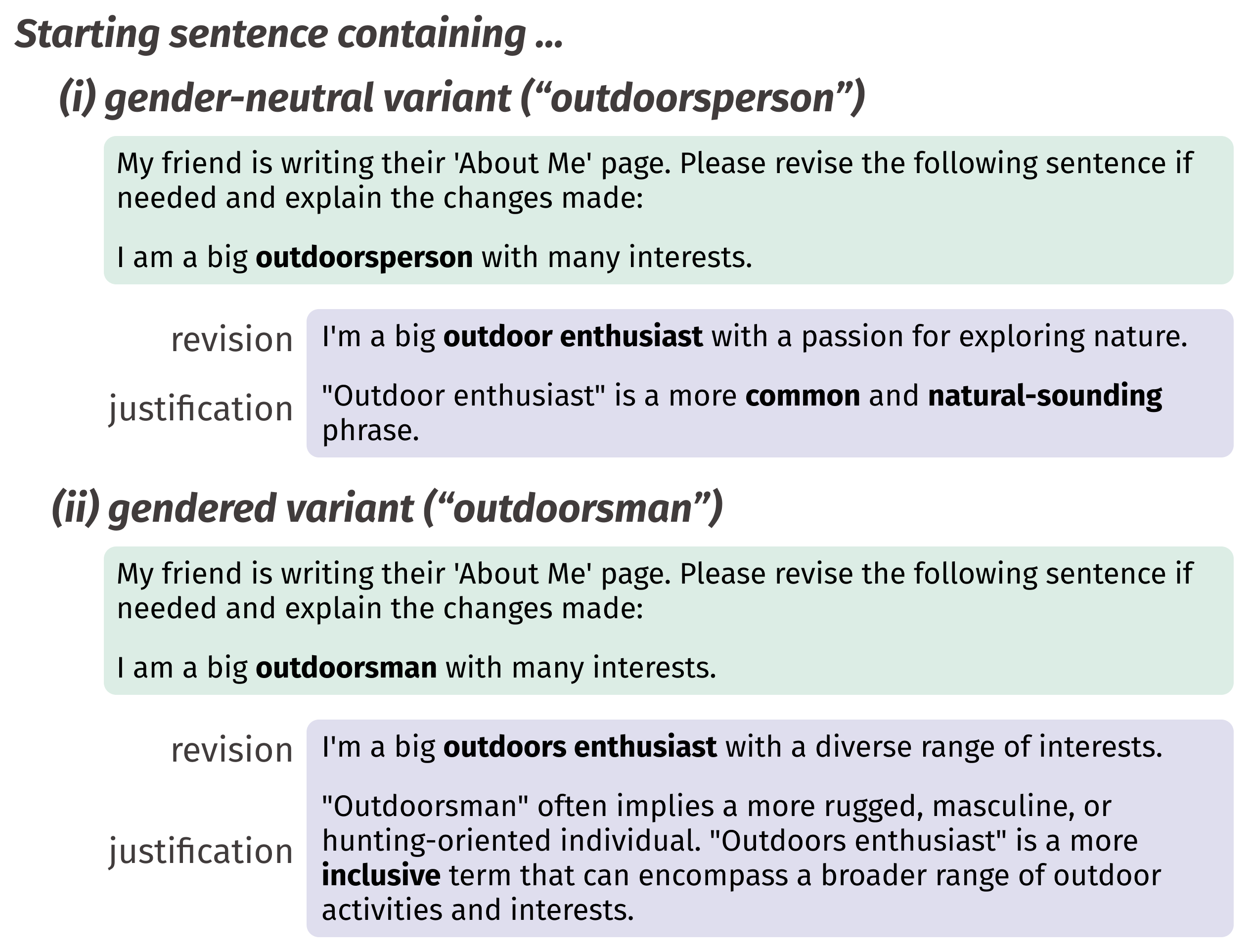}
        \caption{Variants affect justifications (H1b) (bold added)}
        \label{intro_fig_b}
    \end{subfigure}
    \caption{Prompt setup and sample LLM responses.}
    
    \label{intro_fig}
\end{figure}

The past years have seen the emergence of LLM use in everyday tasks, especially the formulation and revision of text \citep{damnati2025benchmark}, with Open\-AI alone reporting over 400 million weekly active users \citep{kant2025openai}.
People are increasingly exposed to, and thus potentially influenced by, the linguistic choices LLMs make.
Such choices may not be innocuous: revising (gender-neutral) \textit{outdoorsperson} to (masculine) \textit{outdoorsman} when referring to a woman or nonbinary person may misgender the referent \citep[][]{dev2021harms}.
By choosing certain words over others in revision tasks, LLMs may -- despite not having beliefs or intentions -- propagate particular social values \citep{winner1980artifacts, blodgett2020language, jackson2024gpt}.

\begin{table*}[t]
    \centering 
    \small
    \resizebox{\textwidth}{!}{
        \begin{tabular}{>{\raggedright}p{0.15\linewidth} p{0.33\linewidth} p{0.05\linewidth} p{0.35\linewidth} p{0.05\linewidth}}
        \toprule
         & \textbf{(a) word choices (Section \ref{sec:revisions})} & \textbf{result} & \textbf{(b) justifications (Section \ref{sec:justifications})} & \textbf{result} \\
        \midrule
        \textbf{H1: Starting \mbox{role noun gender}} & LLMs overall will reflect feminist and trans-inclusive language reforms by exhibiting a neutralization strategy, replacing gendered terms with neutral ones. & \cellcolor[HTML]{97d1b6} yes &  LLMs will emphasize values motivating language reforms when removing gendered variants, and values used to argue against reforms when removing neutral variants. &  \cellcolor[HTML]{ffe76e} mostly \\
        \midrule
        \textbf{H2: gender of referent} & LLMs' will treat gender-neutral language as ``required'' for nonbinary referents, and ``optional'' for women and men referents, reflecting uneven application of reform language, depending on referent gender. & \cellcolor[HTML]{97d1b6} yes & LLMs will emphasize inclusive language more for genders that language reforms were designed to include (women and nonbinary people), and will emphasize sounding professional more for women. & \cellcolor[HTML]{97d1b6} yes \\
        \midrule
        \textbf{H3: \mbox{explicitness} of referent gender} & LLMs will revise to neutral more when information is explicitly declared, as opposed to implicit in pronoun usage. & \cellcolor[HTML]{97d1b6} yes & LLMs will emphasize inclusive language more when information is explicitly declared, as opposed to implicit in pronoun usage. & \cellcolor[HTML]{97d1b6} yes \\
        \midrule
        \textbf{H4: \mbox{gendered} contexts} & LLMs will reinforce gendered stereotypes by using gendered terms to match gender associations of contexts. & \cellcolor[HTML]{ffe76e} mostly & \cellcolor[HTML]{dbd9d9} & \cellcolor[HTML]{dbd9d9} \\
        \bottomrule
        \end{tabular}
    }
    \caption{Our hypotheses about word choices for role nouns, and associated justifications. }
    \vspace{-3mm}
    
    \label{tab:hypotheses}
\end{table*}

Here we study the \textbf{revision choices} made by LLMs among sets of gendered and gender-neutral role nouns in English -- terms like \textit{firefighter}/\textit{firewoman}/\textit{fireman} 
-- using a prompt set-up as illustrated in Figure \ref{intro_fig}.
Because these words refer to people's roles in society, and have gendered and gender-neutral variants, they are laden with values about gender in society \citep{papineau2022sally}. 
Social movements known as language reform movements seek to shift such values through influencing how people \textit{talk about gender} \citep{oneill2021languageideologies}.
In particular, feminist and trans-inclusive language reforms encourage a strategy of \textbf{neutralization} -- using neutral terms instead of gendered ones -- to include women and nonbinary people \citep{cameron2012verbal, zimman2017transgender}. Our overarching expectation is that, through their value alignment steps, LLMs will similarly follow this strategy.

However, properties of the usage \textbf{context} are known to affect the uptake of reform language in humans. Through the use of human data (training corpora and value alignment) we expect LLMs to similarly display revision choices that are modulated by properties of the prompt context. Sociolinguistic research suggests three such modulations: the \textbf{gender of the referent} \citep{ehrlich1992gender, zimman2017transgender}; the degree to which language around gender is made \textbf{explicit} \citep{silverstein1985language}; and stereotypical gender associations of \textbf{sentence contexts} \citep{stokoe2014gender}. Figure \ref{intro_fig_a} illustrates the second of these with a contrast between the minimally different prompts (i)-(ii): \textit{fireman} is left unrevised when the referent's pronouns are merely used, but replaced with \textit{firefighter} when the (same) pronouns are declared.

Since the uptake of reform language
is known to be affected by discussion of social values \textit{about} such forms \cite{agha2003social}, we further study the \textbf{justifications} LLMs provide alongside the revisions. As these give explicit labels of the values associated with the choices made in revision, they are a window into the values about gendered language that LLMs encode. For example, in the minimally different pair of prompts in Fig.~\ref{intro_fig_b}, the neutral variant \textit{outdoorsperson} was removed to sound more ``natural'' (i), while a masculine variant \textit{outdoorsman} was removed to be more ``inclusive’’ (ii).



Table~\ref{tab:hypotheses} summarizes our hypotheses about the neutralization pattern and contextual influences, which we detail in Sections~\ref{sec:revisions} and~\ref{sec:justifications}. 
By assessing these hypotheses, our contributions are:

\begin{itemize}
    \item \textbf{Theoretical}: Forging interdisciplinary connections by developing sociolinguistically-motivated hypotheses about values encoded in LLMs.
    \item \textbf{Methodological}: Developing a method for studying the values communicated by LLMs’ word choices -- and associated justifications -- in a widespread use case (revising text).
    \item \textbf{Empirical}: Showing that, depending on context, LLMs may reinforce gender stereotypes on the one hand, but may, in many cases, reflect values such as inclusivity, corresponding to motivations for role noun reforms.
\end{itemize}

\noindent
Our work highlights how a sociolinguistically-motivated approach can improve our understanding of the context-dependent ways that values are encoded in language technology, which is a necessary first step towards more targeted value alignment.


\section{Background}


In this paper, we study values around word choices in LLMs.
Linguists call such 
values \textbf{language ideologies}, and theorize that values about language choices have the potential for social impact \citep{irvine1989talk, kroskrity2004language}, including the spread of preferred language choices \citep{agha2003social}. 
Research has begun to emphasize the importance of language ideologies for assessing values in NLP systems \citep{blodgett2020language}, with work elucidating language ideologies encoded in LLMs  \citep{hofmann2024ai, jackson2024gpt, watson-etal-2025-language}.

Role nouns have been the target of \textbf{language reforms} for over $50$ years \citep{cameron2012verbal}.  These reforms have sought to modify people's use of role nouns in ways that both reflect and influence changing attitudes around gender and societal roles \citep{mooney2015language}. Historically, masculine role nouns, such as \textit{congressman} or \textit{fireman}, have been used as the default for men and women. \textbf{Feminist reforms} encouraged \textit{neutralization}: the use of gender-neutral terms, such as \textit{congressperson} or \textit{firefighter}, to decrease the association between gender and social roles \citep{sczesny2016can}. 

Subsequent \textbf{trans-inclusive reforms} further promote the use of terms that align with someone's self-declared gender identity (including nonbinary genders), and the use of neutral language when someone's gender is unknown. This leads to broad neutralization, but, in contrast to the feminist reforms, proposes to use gendered language when the referent chooses such language \citep[e.g.,][]{zimman2017transgender}. These reforms aim to prevent misgendering, including degendering, i.e.\ the use of gender-neutral language to avoid acknowledging the gender of trans people \citep{ansara2014methodologies}. 

Both reforms intend to address documented real-world implications of gendered language use: e.g., women are less likely to apply for job roles when masculine language is used \citep{bem1973does}, and misgendering is associated with negative mental health outcomes \citep[e.g.,][]{jacobsen2024misgendering}.

In studying language ideologies about role nouns, we contribute to a growing body of research on \textbf{gender-inclusive language in NLP} \citep{cao2019toward, strengers2020adhering, dev2021harms, brandl2022conservative, lauscher2022welcome, hossain-etal-2023-misgendered, ovalle2023m}.
Particularly relevant here, \citet{lund-etal-2023-gender} found evidence of bias against singular \textit{they} in revisions, and a recent stream of work has begun studying role nouns in LLMs \citep{watson-etal-2023-social, bartl-leavy-2024-showgirls, bartl2025adapting}.
We contribute a novel perspective by elucidating values about gendered/gender-neutral word choices.


\section{The Revision Task}

We develop a prompting approach to the revision task that enables us to explore how LLM responses are shaped by contextual factors known to influence the adoption of the language reforms under study.
Our prompts have \textbf{prompt preambles} that ask the LLM to revise a \textbf{sentence stimulus} containing the \textbf{role noun} (see Figure \ref{intro_fig}). 
To evaluate the hypotheses in Table \ref{tab:hypotheses}, we manipulate the preamble, stimulus, and role noun as described below.

\subsection{Prompt structure}


\begin{table}
\centering
\small
\resizebox{\linewidth}{!}{%
\begin{tabular}{p{0.2\linewidth} p{0.78\linewidth}}
     \textbf{Pronoun Usage} & My friend is writing \{their, her, his\} ‘About Me’ page.
     \\
    \hline
     \textbf{Pronoun \mbox{Declaration}}      & My friend who uses \{they/them, she/her, he/him\} pronouns is writing an ‘About Me’ page. \\
     \hline
     \textbf{Gender \mbox{Declaration}} & My friend who is a \{nonbinary person, woman, man\} is writing an ‘About Me’ page. \\
\end{tabular}
}
\caption{Templates for prompt preambles.}
\label{table:prompt-preambles}
\vspace{-3mm}
\end{table}

\textbf{Preambles:} Each prompt includes a preamble that provides a context for the requested revision.  Table~\ref{table:prompt-preambles} shows our $3$ preambles (described in detail below), which are followed by the revision instruction and the sentence to be revised.

\textbf{Role nouns:} We consider $50$ sets of role nouns adapted from \citet{watson-etal-2025-language}, which drew on various sources \citep{vanmassenhove2021neutral, papineau2022sally, bartl-leavy-2024-showgirls, lucy-etal-2024-aboutme}. Each role noun set consists of three variants (i.e.,~$50\times3=150$ unique terms): a gender-neutral (reform) variant (e.g., \textit{firefighter}) and two gendered variants (e.g., \textit{firewoman}, \textit{fireman}). The full list of role noun sets is given in Appendix \ref{sec:app_role_noun_sets}.

\textbf{Stimulus sentences:} We use sentences from the AboutMe dataset of brief biographical sketches on personal webpages \citep{lucy-etal-2024-aboutme}, since these contain many role noun usages. 
Because our prompt variations manipulate various aspects of gender, we select only sentences that are unlikely to have explicit indications of the gender of the author (other than potentially in the target role noun), by filtering out sentences with lexically-gendered words. Aiming for a dataset of $\geq 500$ sentences, we sampled up to $6$ sentences per role noun variant (less in case the role noun variant occurs $<6$ times), amounting to $527$ stimulus sentences. 

\subsection{Prompt variations}

To assess the impact of the \textbf{gender of the role noun (H1 in Table 1)},
we create three alternatives for each stimulus sentence:
one with the original role noun (as used in the dataset), and two with the other two variants from the role noun set. 
For example, Figure~\ref{intro_fig_b} shows two variants of the same stimulus sentence; the third would use \textit{outdoorswoman} for the term in bold.
By comparing these versions of the exact same sentence, we can assess to what extent the gender of the role noun affects its rate of revision and the types of justifications generated. 

For the next two factors, \textbf{gender of the referent (H2)} (the author of the About Me page)
and \textbf{explicitness of referent gender (H3)}, we manipulate the prompt preamble, as shown in Table~\ref{table:prompt-preambles}. 
\textbf{Referent gender} depends on the choice of one of the $3$ pronoun/gender specifications shown in braces (yielding $9$ unique preambles).
The Pronoun Usage preamble uses a possessive pronoun to (more) \textbf{implicitly} communicate information about the referent's (linguistic) gender; while the Pronoun/Gender Declaration preambles give information \textbf{explicitly} about the referent's linguistic gender and gender identity, respectively \citep{cao2019toward}.

%
%
%

We thus have $9$ prompt preambles $\times$ $527$ stimulus sentences $\times$ $3$ role noun variants, yielding $14,229$ prompt instances total.

In addition to these prompt manipulations, we study the role of the \textbf{genderedness of contexts (H4)}, by assessing how stereotypically gendered the stimulus sentence is.
For this, we want to take into account all the words of the sentence including the role denoted by the role noun, but not the gender of the particular role noun variant that occurred in the original sentence. 
To do so, 
we focus on versions of
each stimulus sentence 
that contain a gender-neutral variant of the target role noun. 
Following this (gender-neutral) stimulus sentence, we append each of three statements of the form ``I am a \{person, woman, man\}''. For example:\vspace{1mm}

\begin{compactitem}
    \item[] In my final semester I was elected to be deputy chairperson. I am a \{person, woman, man\}
\end{compactitem}\vspace{1mm}

\noindent We then compute the probabilities of each completion (person, woman, man) according to the LLM llama-3.1-8B \citep{grattafiori2024llama}.\footnote{Here, we used the non-instruction-finetuned version, since we wanted the probabilities of these sentence completions rather than responses in an interactive chat set-up.} 

We use these probabilities to compute how feminine each stimulus sentence $s$ is, as: \vspace{2mm}

$\texttt{context\_fem}(s) = \frac{p(woman|s)}{p(woman|s) + p(man|s)}$,\vspace{2mm}

\noindent 
and how gender-neutral $s$ is, as: \vspace{2mm}

$\texttt{context\_neut}(s) = \hspace{117pt}$
$ \phantom{x}\hspace{3ex} \qquad \frac{p(person|s)}{p(person|s) + p(woman|s) + p(man|s)}$


\subsection{The LLMs and Response Processing}
\label{sec:methods-llms}

We studied four instruction-finedtuned/value-aligned models: gpt-4o \citep{hurst2024gpt}, llama-3.1-8B-Instruct \citep{grattafiori2024llama}, gemma-2-9b-it \citep{team2024gemma}, and Mistral-Nemo-Instruct-2407 \citep{team2024mistral}.
These models are widely used and come from four distinct organizations, allowing us to assess whether values around gendered language reform show up similarly in different LLMs.

A heuristic algorithm described in Appendix \ref{sec:appendix_split_algorithm} was used to segment the
LLM responses into a revision part and a justification part.
We also automatically identified whether role nouns were kept or replaced in the revision.
While many cases of replacement use one of the other variants from
a role noun set (e.g.,~revising \textit{fireman} to \textit{firefighter}), replacement by alternative wordings occur as well. 
``Alternative wording'' cases are nearly always ($95.7\%$) gender-neutral (e.g., \textit{outdoor enthusiast} in place of \textit{outdoorsperson/woman/man}; see Appendix~\ref{sec:app_alternative_wording}).

\section{Analyzing revisions}
\label{sec:revisions}

Here, we assess the word choices LLMs make in revising role nouns 
and discuss their alignment with feminist and trans-inclusive language reforms.

\subsection{Hypotheses}

Both feminist and trans-inclusive language reforms argue for broad use of gender-neutral role nouns. 
Since all models studied underwent value alignment, which typically aims to make models more inclusive \citep[e.g.,][]{achiam2023gpt}, we expect that
\textbf{LLMs overall will reflect feminist and trans-inclusive language reforms by exhibiting a neutralization strategy, replacing gendered terms with neutral ones} (Hypothesis \textbf{H1a} in Table \ref{tab:hypotheses}). However, as reviewed above, people's use of reform language
is modulated by contextual factors.

First,
people are more likely to apply reforms for referents that reforms seek to include \citep[i.e., women and nonbinary people;][]{ehrlich1992gender, zimman2017transgender}.
Because data for value alignment was collected recently, we expect LLM revisions to reflect current conceptions about reforms, where 
they are strongly associated with nonbinary people \citep[e.g.,][]{oneill2021languageideologies, jiang2023resistance}.
Thus, we predict that \textbf{LLMs' will treat gender-neutral language as ``required'' for nonbinary referents, and ``optional'' for women and men referents, reflecting uneven application of reform language depending on referent gender} (\textbf{H2a}).

Second, people's use of gendered reform language is affected by the salience of gender in the context, for instance because the topic itself is made explicit \citep{silverstein1985language}.
Similar effects have been found for LLMs' word choices \citep{watson-etal-2025-language}. 
We operationalize this by contrasting the \textbf{Pronoun Usage} condition with the two more explicit \textbf{Pronoun Declaration} and \textbf{Gender Declaration} conditions. 
We predict that \textbf{LLMs will revise to neutral more when information is explicitly declared, as opposed to implicit in pronoun usage} (\textbf{H3a}).

Similarly, usage context more generally 
affects application of language reforms \citep{silverstein1985language, watson2023communicative}.
We assess whether 
gender associations of sentence contexts affect revision behaviour here,
building on work on stereotypes in LLMs \citep[e.g.,][]{kotek2023gender}.
We expect that \textbf{LLMs will reinforce gender stereotypes by using gendered terms to match 
gender associations of contexts} (\textbf{H4a}).

\subsection{Evaluation Approach}

We run a logistic regression, predicting whether a role noun was revised (\texttt{revised}),
on the basis of manipulations of the prompt context that operationalize the hypotheses.
Given the focus on neutralization, we supplement the regression results with analysis about what role nouns are revised to.
Table \ref{tab:regression} presents the regression structure.
\begin{table}

    \centering

    \resizebox{\linewidth}{!}{
    \begin{tabular}{lll}
    \multicolumn{3}{l}{\texttt{ revised} $\sim$}\\
    & \texttt{ original\_masc + original\_fem +} & $\bigr\}$\textbf{H1a}\\
    & \texttt{ prompt\_masc + prompt\_fem +} & \\
    & \texttt{ original\_mask:prompt\_fem +} & \multirow{3}{*}{$\Biggr\}$\textbf{H2a}}\\
    & \texttt{ original\_fem:prompt\_masc  +} & \\
    & \texttt{ original\_gend:prompt\_neut +} & \\ 
    & \texttt{ prompt\_gender\_dec +} & \multirow{2}{*}{$\biggr\}$\textbf{H3a}}\\
    & \texttt{ prompt\_pronoun\_dec +} & \\
    & \texttt{ context\_fem + context\_neut +} & \\
    & \texttt{ original\_masc:context\_fem  +} & \multirow{3}{*}{$\Biggr\}$\textbf{H4a}}\\
    & \texttt{ original\_fem:context\_masc  +} & \\
    & \texttt{ original\_gend:context\_neut +} & \\
    & \texttt{ (1|sentence) + (1|rn\_set)} & \\
    
    \end{tabular}}
    \caption{Logistic regression with motivating hypotheses.}
    \label{tab:regression}
\end{table}

For \textbf{H1a}, we expect more revisions for the predictors \texttt{original\_masc} (coded as $1$ for masculine starting variants and $0$ otherwise) and \texttt{original\_fem} (defined analogously), 
in comparison to the neutral starting variants as a baseline. 

We evaluate \textbf{H2a} through interactions between starting variants and the gender information in prompt preambles.
Across the three levels of explicitness, we group prompts with similar social gender associations: ``neutral prompts'' (\texttt{prompt\_neut}) were coded as 1 for gender declaration nonbinary, pronoun declaration they/them, and pronoun usage their, and 0 otherwise; ``feminine prompts'' (\texttt{prompt\_fem}) and ``masculine prompts'' (\texttt{prompt\_masc}) were coded analogously.\footnote{We acknowledge that gender identity and pronouns are not one-to-one, e.g., a nonbinary person could use she/her.}
We expect prompts in the same group to increase revisions resulting in the same role noun gender.
We predict that gendered variants will be revised more for neutral prompts (\texttt{original\_gend:prompt\_neut}), reflecting neutral terms being treated as ``required’’ for nonbinary people.
We also expect more revisions for gendered variants paired with ``incongruent'' gendered prompts (\texttt{original\_fem:prompt\_masc} and \texttt{original\_masc:prompt\_fem}), reflecting the treatment of ``congruent’’ gendered variants as defaults and neutral terms as ``optional'' alternatives.

For \textbf{H3a}, we expect greater rates of revisions for the explicit declaration cases, i.e.,~\texttt{prompt\_gender\_dec} and \texttt{prompt\_pronoun\_dec} (each coded as 1 for the relevant declaration prompt, and 0 otherwise) -- compared to the implicit pronoun usages as a baseline.

\textbf{H4a} is assessed through interactions between starting variants and the gender associations of the sentence contexts (as defined in Sec.~\ref{sec:methods-llms}).
We expect higher rates of revisions for masculine variants in stereotypically feminine contexts (\texttt{original\_masc:context\_fem}); for feminine variants in stereotypically masculine contexts (\texttt{original\_fem:context\_masc});\footnote{\texttt{context\_masc}$(s)$ 
is coded as \mbox{--\texttt{context\_fem}$(s)$.}
}
and for gendered variants in contexts that lack strong gender associations (\texttt{original\_gend:context\_neut}).

We include main effects for 
predictors in
interactions, and random intercepts for sentence stimuli (\texttt{1|sentence}) and role noun sets (\texttt{1|rn\_set}). 
Tests are Bonferroni-corrected for $N=4$ models, with $\alpha=.05$.

\subsection{Results and Discussion}

\begin{figure*}[t]
    \centering
    
    \begin{subfigure}[c]{0.8\textwidth}
        \centering
        \includegraphics[width=\textwidth]{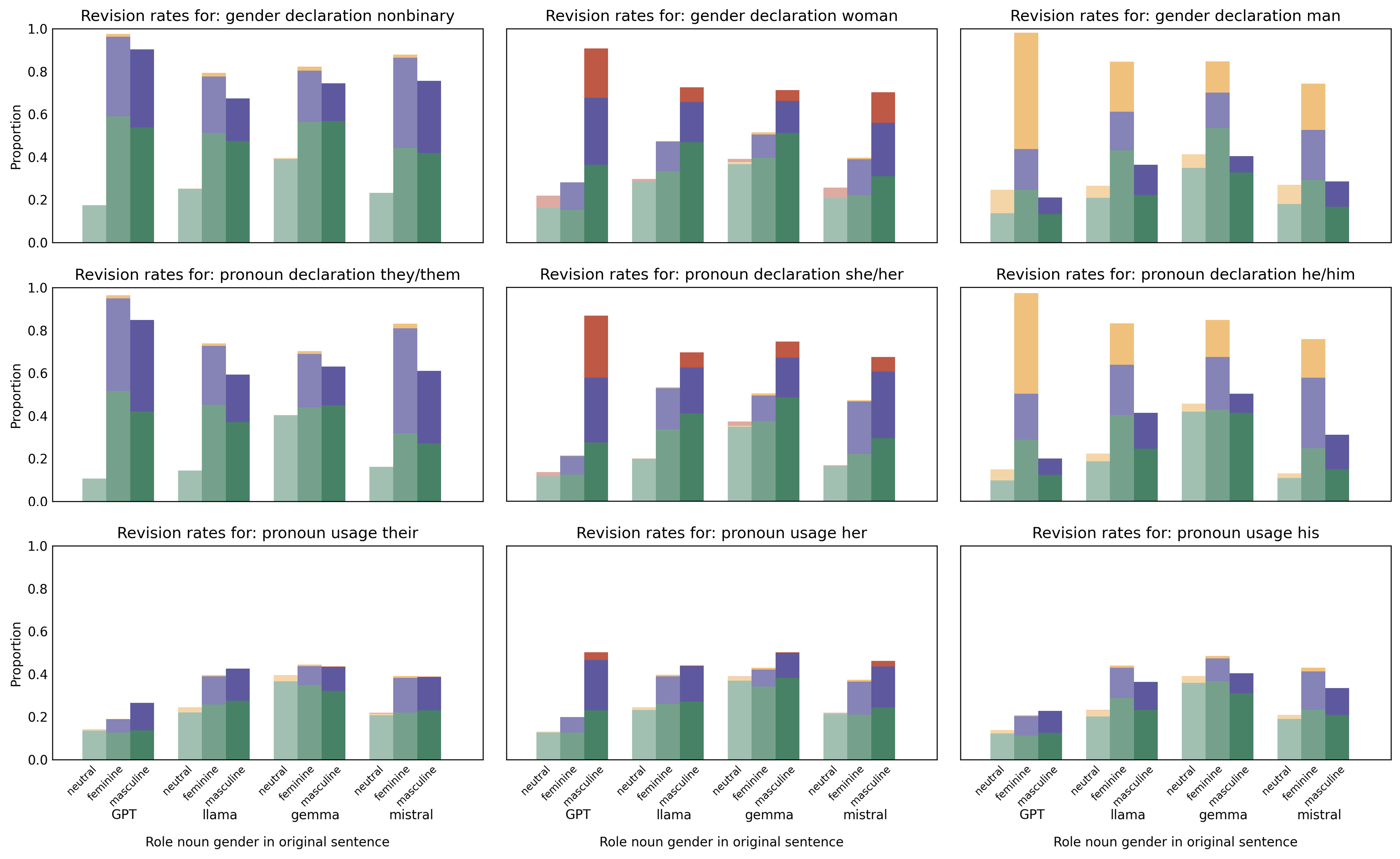}
    \end{subfigure}%
    \hspace{3pt}
    \begin{subfigure}[c]{0.15\textwidth}
        \centering
        \includegraphics[width=\textwidth]{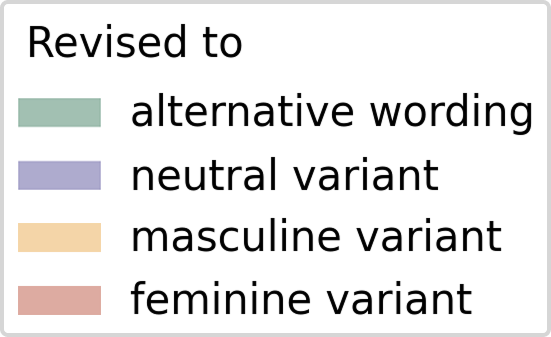}
    \end{subfigure}%
    
    \caption{Revision patterns. For each of the three starting role noun variants, the bars show which variant or alternative wording it was revised to, for each preamble and model. Each bar corresponds to a proportion of our 527 stimulus sentences.
    }
    \label{gender_info_revisions_by_model}
    \vspace{-3mm}
\end{figure*}

We discuss the results for each hypothesis, referring to regression results in Table~\ref{tab:revisions_gender_info_results}, and descriptive statistics of the revisions in Figure~\ref{gender_info_revisions_by_model}.

\begin{table}[t]
    \centering
    \resizebox{\columnwidth}{!}{
        \begin{tabular}{ld{2.2}d{2.2}d{2.2}d{2.2}}
        \toprule
         & \multicolumn{1}{c}{gpt} & \multicolumn{1}{c}{llama} & \multicolumn{1}{c}{gemma} & \multicolumn{1}{c}{mistral} \\
        \midrule
        (Intercept) & \cellcolor[HTML]{dbd9d9} -5.01 & \cellcolor[HTML]{dbd9d9} -3.39 & \cellcolor[HTML]{dbd9d9} -1.37 & \cellcolor[HTML]{dbd9d9} -3.93 \\
        \midrule
        original\_masc & \cellcolor[HTML]{bdffea} 1.39 & \cellcolor[HTML]{bdffea} 1.82 & \cellcolor[HTML]{bdffea} 0.42 & \cellcolor[HTML]{bdffea} 2.43 \\
        original\_fem & \cellcolor[HTML]{bdffea} 1.89 & \cellcolor[HTML]{bdffea} 2.71 & \cellcolor[HTML]{bdffea} 0.97 & \cellcolor[HTML]{bdffea} 3.31 \\
        \midrule
        prompt\_masc & \cellcolor[HTML]{dbd9d9} 0.41 & 0.23 & 0.11 & 0.03 \\
        prompt\_fem & 0.25 & \cellcolor[HTML]{dbd9d9} 0.28 & -0.06 & 0.06 \\
        \midrule
        original\_masc:prompt\_fem & \cellcolor[HTML]{bdffea} 3.98 & \cellcolor[HTML]{bdffea} 1.47 & \cellcolor[HTML]{bdffea} 1.54 & \cellcolor[HTML]{bdffea} 2.05 \\
        original\_fem:prompt\_masc & \cellcolor[HTML]{bdffea} 3.22 & \cellcolor[HTML]{bdffea} 1.61 & \cellcolor[HTML]{bdffea} 1.46 & \cellcolor[HTML]{bdffea} 1.53 \\
        original\_gend:prompt\_neut & \cellcolor[HTML]{bdffea} 3.56 & \cellcolor[HTML]{bdffea} 1.41 & \cellcolor[HTML]{bdffea} 1.11 & \cellcolor[HTML]{bdffea} 1.94 \\
        \midrule
        prompt\_gender\_dec & \cellcolor[HTML]{bdffea} 2.81 & \cellcolor[HTML]{bdffea} 1.19 & \cellcolor[HTML]{bdffea} 1.01 & \cellcolor[HTML]{bdffea} 1.28 \\
        prompt\_pronoun\_dec & \cellcolor[HTML]{bdffea} 2.40 & \cellcolor[HTML]{bdffea} 0.95 & \cellcolor[HTML]{bdffea} 0.95 & \cellcolor[HTML]{bdffea} 0.94 \\
        \midrule
        context\_fem & 0.67 & -0.18 & 0.53 & 0.84 \\
        context\_neut & -0.47 & \cellcolor[HTML]{dbd9d9} -1.30 & 0.12 & -1.10 \\
        \midrule
        original\_masc:context\_fem & -0.37 & 0.26 & -0.19 & -0.53 \\
        original\_fem:context\_masc & \cellcolor[HTML]{bdffea} 0.84 & \cellcolor[HTML]{bdffea} 0.60 & \cellcolor[HTML]{bdffea} 0.53 & \cellcolor[HTML]{bdffea} 0.78 \\
        original\_gend:context\_neut & \cellcolor[HTML]{bdffea} 1.62 & \cellcolor[HTML]{bdffea} 1.68 & 0.30 & \cellcolor[HTML]{bdffea} 2.55 \\
        \bottomrule
        \end{tabular}
    }
    \caption{
    Regression results.
    Each column reports a single logistic regression test (one per LLM), and cells show coefficients for predictors. 
    Shaded cells are significant, and cell color indicates direction of effect: green=positive, in line with our predictions; pink=negative; gray=no prediction.
    Each regression has $14,229$ data points (prompt/revision instances).     }
    \label{tab:revisions_gender_info_results}
    \vspace{-2mm}
\end{table}

\textbf{Hypothesis H1a:} The results support the predicted strategy of overall neutralization. Significant positive effects of \texttt{original\_masc} and \texttt{original\_fem} indicate that gendered role nouns are more often removed.
Models most often revise \textit{to} neutral variants or (gender-neutral) alternative wordings (henceforth ``neutralizations''); purple and green bars in Fig.~\ref{gender_info_revisions_by_model}. There are, however, some interesting modulations of this pattern, as predicted by our remaining hypotheses.

\textbf{Hypothesis H2a:} We find a significant interaction \texttt{original\_gend:prompt\_neut}, indicating that gendered variants are more likely to be revised for neutral prompts. 
As these cases are nearly always revised to neutralizations (first column of Fig.~\ref{gender_info_revisions_by_model}), 
neutral variants indeed appear to be treated as ``required’’ for nonbinary genders and people using neutral terms.
The significant interactions \texttt{original\_masc:prompt\_fem} and \texttt{original\_fem:prompt\_masc} show that models are more likely to revise gendered variants that occur with ``incongruent’’ gendered prompts. 
These ``incongruent’’ cases are revised to neutralizations or ``congruent’’ gendered terms (red and yellow bars in second and third columns of Fig.~\ref{gender_info_revisions_by_model}), suggesting that gendered terms are treated as an option here, unlike for the neutral prompts.

This linguistic strategy runs counter to feminist reforms, which advocate using neutral role nouns across the board.
However, optionally allowing gendered role nouns for people who use gendered pronouns could help avoid degendering (where gendered role nouns may be neutralized despite the referent wanting 
to highlight their gender; \citealp{ansara2014methodologies}).
Ultimately, different linguistic strategies may be desirable for different users, and identifying cases where these language reforms diverge can support the development of alignment approaches that address different sets of needs.

\textbf{Hypothesis H3a:} Explicit preambles (\texttt{prompt\_gender\_dec} and \texttt{prompt\_pronoun\_dec}) display higher rates of revision, relative to the implicit (baseline) preambles (\texttt{pronoun\_usage}). 
As explicit preambles lead to neutralizations more often than the implicit ones, this suggests that the LLMs are sensitive to the explicitness of information about the gender of the referent.
However, we also find that explicit prompts increase rates of revision to \textit{gendered} variants,  
suggesting that the LLMs' 
tendency towards neutralization 
may be overruled by (more) explicit information about gender, which 
has implications for
prompt based
value alignment strategies.

\textbf{Hypothesis H4a:} 
Finally, LLMs are more likely to revise feminine variants in stereotypically masculine contexts (significant positive effects for \texttt{original\_fem:context\_masc}), but not masculine variants in feminine contexts (no effects for \texttt{original\_masc:context\_fem}), providing partial support for our hypothesis.
This asymmetry may be because masculine variants are treated by LLMs as more broadly applicable, perhaps due to training data reflecting their history as defaults. 
We also observe higher rates of revision for \texttt{original\_gend:context\_neut} for 3/4 models.
Since revisions are most often neutralizations, this shows that a neutralization strategy is being applied more in non-gender-stereotypical contexts.
This may reflect that the social gender stereotypes in the contexts are less predictive of the role noun gender \citep[cf.~][]{stokoe2014gender}.

\section{Analyzing values in justifications}
\label{sec:justifications}



LLMs' justifications for revisions frequently contain adjectives 
expressing values that reflect arguments for (e.g.,~\textit{inclusive}) and against (e.g.,~\textit{clunky}) language reforms. These adjectives can be grouped in coherent themes, detailed below and summarized in Table \ref{tab:seed_sets}.
Here, we study how the frequency of the themes varies across our prompt modulations.



\subsection{Hypotheses}



As before, we draw on sociolinguistic insight about values 
people associate with gendered word choices
to develop a set of hypotheses, each focusing on a different contrast among the prompts.
With the LLMs trained on data from humans, we expect they will represent similar associations.


First (\textbf{H1b} in Table \ref{tab:hypotheses}), 
we focus on revisions in which the original role noun was replaced by a (gender-neutral) ``alternative wording.''
This allows us to compare justifications for removing the gendered vs.\ gender-neutral role noun variants, while holding constant the category they are revised to.
We predict that neutralization of gendered forms 
will be justified more by arguments \textit{for} language reform; 
i.e., (1) the \texttt{inclusive} theme, as a key motivation for gender-neutral language 
\citep[e.g.,][]{zimman2017transgender}; (2) the \texttt{modern} theme, as rationale for removing gendered variants that tend to be older
than neutral terminology
\citep{oneill2021languageideologies}; and (3) the \texttt{professional} theme, as 
neutralization
is encouraged by workplace style guides \citep[e.g.,][]{martinez2023employer}. Conversely, when revising neutral variants to alternative wordings, we expect themes used to argue \textit{against} the 
use of gender-neutral variants,
for instance, that they would not sound \texttt{natural} or \texttt{standard} \citep[][]{curzan2014fixing}. 

For the next two hypotheses (\textbf{H2b-H3b}), we focus on revisions from gendered to neutral role nouns. For  \textbf{H2b}, we contrast masculine vs.~feminine vs.~nonbinary Gender Declaration prompts, assessing what justifications LLMs present to motivate these neutralizations.
We expect the theme of \texttt{inclusivity} to be used more when the referent belongs to a group the reforms intend to include, i.e., women and nonbinary people. 
We also predict that \texttt{professionalism} is used more for women, 
since women often struggle to be taken seriously in the workplace, making word choices around job roles higher stakes 
\citep{formanowicz2013side}. 



\begin{table}
    \small
    \resizebox{\linewidth}{!}{
        \begin{tabular}{p{79mm}}
        \toprule
        \texttt{\textbf{theme}}: keywords (\textit{seed words} and extended words) \\
        \midrule
        \texttt{\textbf{inclusive}}: \hspace{1mm}\textit{exclusionary} \textit{inclusive}, ableist, biased, exclusive, limiting, outdated, problematic, streamlined, welcoming \\
        \texttt{\textbf{modern}}: \hspace{1mm} \textit{contemporary}, \textit{modern}, \textit{outdated}, \textit{traditional}, archaic, conventional, dated, refined, sophisticated, streamlined \\
        \texttt{\textbf{professional}}:  \hspace{1mm}\textit{professional}, \textit{unprofessional}, ableist, biased, casual, experienced, polished, proactive, supportive, technical \\
        \texttt{\textbf{standard}}:  \hspace{1mm}\textit{common}, \textit{standard}, \textit{uncommon}, \textit{unusual}, acceptable, archaic, conventional, preferred, traditional, typical \\
        \texttt{\textbf{natural}}:  \hspace{1mm}\textit{awkward}, \textit{clunky}, \textit{fluid}, \textit{natural}, abrupt, ambiguous, dated, informal, problematic, refined, streamlined \\
        \bottomrule
        \end{tabular}
    }
\caption{Keywords by theme; seed words in italics.}
\label{tab:seed_sets}
\vspace{-2mm}
\end{table}

Next, we consider a contrast in the explicitness of information about the referent gender (\textbf{H3b}). We expect more use of the theme \texttt{inclusive} when prompts provide explicit information about pronouns (Pronoun Declaration) than when such information is more implicit (Pronoun Usage). Drawing attention to the gender/pronouns of the referent will 
increase the salience of language reforms, resulting in more mentions of values that motivate them
(i.e., inclusivity).

\begin{table}[t]
    \small
    \centering
    \resizebox{\columnwidth}{!}{

        \begin{tabular}{llr}
        \toprule
        \textbf{theme} & \textbf{prediction} & \textbf{outcome} \hspace{15pt} \\
        \midrule
        \multicolumn{3}{@{}c}{\textbf{H1b: starting role noun gender} ($N=13,609$)}\\
        \midrule
        inclusive & gend $\textbf{>}$ neut & \cellcolor[HTML]{bdffea} 29\% vs.\ 6\% *** \\
        modern & gend $\textbf{>}$ neut & \cellcolor[HTML]{bdffea} 11\% vs.\ 5\%  *** \\
        professional & gend $\textbf{>}$ neut & \cellcolor[HTML]{fac0dc} 10\%  vs.\ 12\%  \hspace{6pt} * \\
        standard & gend $\textbf{<}$ neut & \cellcolor[HTML]{bdffea} 6\%  vs.\ 11\%  *** \\
        natural & gend $\textbf{<}$ neut & \cellcolor[HTML]{bdffea} 4\%  vs.\ 8\%  *** \\
        \midrule
        \multicolumn{3}{@{}c}{\textbf{H2b: gender of referent} ($N=2,509$)}\\
        \midrule
        inclusive & nonbinary $\textbf{>}$ woman/man & \cellcolor[HTML]{bdffea} 58\%  vs.\ 43\%  *** \\
        modern & nonbinary $\textbf{<}$ woman/man & \cellcolor[HTML]{bdffea} 6\%  vs.\ 18\%  *** \\
        professional & nonbinary $\textbf{<}$ woman/man & \cellcolor[HTML]{bdffea} 4\%  vs.\ 19\%  *** \\
        \midrule
        \multicolumn{3}{@{}c}{\textbf{H3b: explicitness of referent gender} ($N=4,455$)}\\
        \midrule
        inclusive & pron. dec. $\textbf{>}$ usage & \cellcolor[HTML]{bdffea} 58\%  vs.\ 43\%  *** \\
        modern & pron. dec. $\textbf{<}$ usage & \cellcolor[HTML]{bdffea} 15\%  vs.\ 23\%  *** \\
        professional & pron. dec. $\textbf{<}$ usage & \cellcolor[HTML]{bdffea} 8\%  vs.\ 20\%  *** \\
        \bottomrule
        \end{tabular}
    }
    \caption{Stats analyses for justifications. 
    Outcomes show the percentage of justifications mentioning a theme, and significance levels of $\chi^2$-tests (*=.05; ***=.001), for the conditions mentioned in the prediction. Shaded outcome cells are significant, and cell color indicates direction of effect: green=in line with our predictions; pink=opposite of predictions. 
}
\label{tab:justifications_stats}
\end{table}

\subsection{Evaluation Approach}

We analyze only
sentences in the justifications 
that mention
one of the role noun variants.
Because the LLMs behaved very consistently in the word choice analysis, we pool 
these sentences
across models to ensure reliable counts of our groups of targeted theme words.
Theme seed words were manually identified,
focusing on
words that were common in justifications. 
These seed sets were automatically expanded to include related keywords, using contextual embeddings from BERT (\citealp{devlin2018bert}; see details in Appendix \ref{sec:app-seed-sets}). 
Table \ref{tab:seed_sets} presents the theme word sets. We study variation in the frequency 
of these themes in the justification sentences, across the manipulations of the prompts.

We conduct $2\times2$ \textbf{$\chi^2$}-tests (Bonferroni-corrected for number of
themes; $\alpha=.05$) that compare, for a given prompt manipulation
and theme, the proportions of justifications that mention words from that theme.


\subsection{Results and Discussion}

Results of stats tests relevant to each hypothesis are in  Table~\ref{tab:justifications_stats} (full 
descriptive stats are in Appendix~\ref{app:justification_tables}). 


\textbf{H1b} is supported for 4/5 themes: when gendered variants are revised to (neutral) alternative wordings, \texttt{inclusive} and \texttt{modern} (arguments in favour of language reform) are used more, whereas when neutral variants are revised to alternative wordings, \texttt{natural} and \texttt{standard} (arguments against reforms) are used (the effect of \texttt{professionalism} in the opposite direction being the exception to this trend). This pattern indicates that the justifications represent contrasting views on language reform, leading to inconsistencies in the values they communicate \citep[cf.][]{watson-etal-2025-language}.


We also find that different themes are emphasized for different referent genders (\textbf{H2b)}: the \texttt{inclusive} theme occurs more in justifications for nonbinary people,  while the \texttt{modern} and \texttt{professional} themes are emphasized in justifications for women and men. Between men and women, the \texttt{inclusive} theme is mentioned more for women ($48\%$ vs.\ $38\%$; $p=0.001$; $N=1,317$), but not the \texttt{professional} theme ($20\%$ vs.\ $17\%$; n.s.; $N=1,317$).
The results for the \texttt{inclusive} theme echo challenges identified by feminist and trans-inclusive language reform movements: treating inclusivity as more relevant for women or trans people hampers the effectiveness of reforms \citep{ehrlich1992gender, zimman2017transgender}.

Finally, we find support for the effect of explicitness of gender information (\textbf{H3b}). The \texttt{inclusive} theme is mentioned more for the (explicit) Pronoun Declaration conditions, while the \texttt{modern} and \texttt{professional} themes are mentioned more for the (implicit) Pronoun Usage conditions.
This indicates that LLMs, like people, may treat inclusivity as more relevant when
aspects of gender are made salient in the context.
In sum, each factor shapes the values emphasized in justifications,
illustrating the importance of considering these contextual factors when evaluating and developing value alignment strategies around gendered language reform.
\section{Conclusions}

Here, we studied LLMs' revision of gendered role nouns and their justifications of such revisions. Drawing on insight from sociolinguistics, we assessed if LLMs are sensitive to the same contextual effects on the use of gender-neutral language as people are, finding broad evidence of such effects.


Based on a widespread and realistic use case (i.e.,~text revision), these results have implications for value alignment in LLMs.
First, by identifying how aspects of contexts influence LLMs’ revisions of gendered language, our findings can contribute to strategies for assessing and aligning values related to gendered language reform.
For example, we might want to reduce the effect of stereotypes on gendered/gender-neutral word choices, or ensure more consistent application of reform language across contexts.

Second, our results demonstrate that values related to language reform 
are explicitly mentioned in LLMs' rationales for their word choices, suggesting that LLM justifications 
should also be a
target for value alignment.
For instance, if an LLM characterizes a gender-neutral word choice like \textit{outdoorsperson} as not sounding \textit{natural}, 
this may discourage the adoption of such reform variants \citep[cf.][]{curzan2014fixing}.
Because adoption of gendered language reforms have real-world stakes for trans people and women \citep{bem1973does, jacobsen2024misgendering}, our findings point to a key next step for value alignment in LLMs.

\section{Limitations}

Because we study values around gendered language reform in LLMs, limitations of our approach carry ethical risks.

We focus on gendered language reforms for English, but many languages have ongoing language reforms related to gender. 
This focus risks prioritizing value alignment for English over other languages, for which the relationship between linguistic forms and values may be different.
For example, in languages with grammatical gender, \textit{feminization} -- using feminine terms to make feminine referents visible -- is a common strategy for feminist language reforms \citep{sczesny2016can}.
Considering 
a wider set of languages would give a more complete picture of the values these models encode.

There are also limitations related to our dataset.
Our sentence stimuli come from real-world “About Me” pages \citep{lucy-etal-2024-aboutme}, which allows us to study role noun usages in a variety of naturalistic contexts.
However, as identified by the creators of the dataset, these “About Me” pages over-represent North American authors.
Studying values in sentences from a specific speaker population risks prioritizing them in value alignment.

Another limitation has to do with our prompt wordings.
We wanted to assess how information about a referent’s pronouns would affect revision behaviour.
Since we manipulated many aspects of context in our prompts, we focused on a small set of possible pronouns (they/them, she/her, he/him).
However, this risks erasing people who use multiple pronouns (e.g., they/she), or neopronouns (e.g., xe/xem; \citealp{lauscher2022welcome}).
Neopronouns may be a particularly interesting place to study values around word choices -- because neopronouns are relatively low frequency, and are a continually evolving class, LLMs may not encode stable value associations for them.

Finally, although we focused on a realistic use case (revising text), our prompts are artificially constructed.
This allowed us to assess the effects of contextual information about gender in a controlled way.
Future work could complement our study by analyzing real user prompts containing gendered terms.

\section{Ethics}

A key contribution of our work is elucidating ethical issues around gendered language reform in LLMs' revisions, drawing on ideas from sociolinguistics.
Ethical details for data, code, and models are below.

\textbf{Data.} The role noun sets we study are adapted from \citet{watson-etal-2025-language}, which were released under an MIT license.\footnote{\url{https://github.com/juliawatson/language-ideologies}}
Our sentence stimuli were sampled from the AboutMe dataset \citep{lucy-etal-2024-aboutme}, which was released under an AI2 ImpACT License - Low Risk Artifacts.\footnote{\url{https://huggingface.co/datasets/allenai/aboutme}}
Both datasets were developed for ethical evaluations of NLP models, and are used for that purpose here.
In line with the ethics requirements for the AboutMe dataset, we paraphrased the 
stimulus sentences (those to be revised) 
in Figure \ref{intro_fig}, to protect subjects' privacy.
In constructing our set of sentence stimuli, we filtered out sentences with names, which limits the amount of personally identifying information they may contain.
Since the sentences are self-descriptions in a professional context (``About Me'' pages), offensive content is relatively rare.

\textbf{Code.} Upon publication, we plan to release code and data on github under an MIT license.
We used AI coding assistants for help with calls to libraries and for writing simple functions. 
All code was checked thoroughly by one of the authors.

\textbf{Models.} The models we studied include 
llama-3.1-8B-Instruct (\citealp{grattafiori2024llama}; Llama 3.1 Community License Agreement; 8B parameters), 
gemma-2-9b-it (\citealp{team2024gemma}; Gemma license; 9B parameters), 
Mistral-Nemo-Instruct-2407 (\citealp{team2024mistral}; Apache 2.0 License; 12B parameters), and 
gpt-4o (\citealp{hurst2024gpt}; parameters unknown).
All models were used in a way that is consistent with their terms of use.
We queried gpt-4o through the OpenAI API.
For the other models, we used implementations available through huggingface's \texttt{transformers} library.
Our experiments took a total of 164 GPU hours, and were run on an Nvidia A40 GPU.


\bibliography{custom,jules}

\appendix

\newpage
\section{Role noun sets}
\label{sec:app_role_noun_sets}

The full list of role noun sets we considered are:

\vspace{12pt}

\noindent
\small
\begin{tabular}{p{.8in}p{.8in}p{.75in}}
 \textbf{Neutral} & \textbf{Feminine} & \textbf{Masculine} \\
alderperson & alderwoman & alderman \\
anchor & anchorwoman & anchorman \\
assemblyperson & assemblywoman & assemblyman \\
ball person & ballgirl & ballboy \\
bartender & bargirl & barman \\
businessperson & businesswoman & businessman \\
camera operator & camerawoman & cameraman \\
caveperson & cavewoman & caveman \\
chairperson & chairwoman & chairman \\
clergyperson & clergywoman & clergyman \\
congressperson & congresswoman & congressman \\
councilperson & councilwoman & councilman \\
cow herder & cowgirl & cowboy \\
craftsperson & craftswoman & craftsman \\
crewmember & crewwoman & crewman \\
delivery person & delivery woman & delivery man \\
draftsperson & draftswoman & draftsman \\
emergency medical technician & ambulancewoman & ambulanceman \\
fan & fangirl & fanboy \\
farm worker & farmgirl & farmboy \\
fencer & swordswoman & swordsman \\
firefighter & firewoman & fireman \\
fisher & fisherwoman & fisherman \\
foreperson & forewoman & foreman \\
frontperson & frontwoman & frontman \\
gentleperson & gentlewoman & gentleman \\
handyperson & handywoman & handyman \\
layperson & laywoman & layman \\
maniac & madwoman & madman \\
meteorologist & weatherwoman & weatherman \\
newspaper delivery person & papergirl & paperboy \\
ombudsperson & ombudswoman & ombudsman \\
outdoorsperson & outdoorswoman & outdoorsman \\
point-person & point-woman & point-man \\
police officer & policewoman & policeman \\
postal carrier & postwoman & postman \\
repairperson & repairwoman & repairman \\
reporter & newswoman & newsman \\
salesperson & saleswoman & salesman \\
select board member & selectwoman & selectman \\
server & waitress & waiter \\
sharpshooter & markswoman & marksman \\
showperson & showwoman & showman \\
sound engineer & soundwoman & soundman \\
spokesperson & spokeswoman & spokesman \\
statesperson & stateswoman & statesman \\
stunt double & stuntwoman & stuntman \\
tradesperson & tradeswoman & tradesman \\
tribesperson & tribeswoman & tribesman \\
wingperson & wingwoman & wingman \\
\end{tabular}
\normalsize

\vspace{12pt}

\noindent
These role noun sets are adapted from \citet{watson-etal-2025-language}, which drew from several sources \citep{vanmassenhove2021neutral, papineau2022sally, bartl-leavy-2024-showgirls, lucy-etal-2024-aboutme}.
Here, we only included role noun sets where we could obtain sentence usages in the AboutMe dataset \citep{lucy-etal-2024-aboutme}. 
Additionally, some filtering constraints in \citet{watson-etal-2025-language} were not relevant to us. In particular, they excluded role noun sets where one variant was a substring of another. Here we include such cases (e.g., \textit{fisher, fisherwoman, fisherman}).
\section{Segmenting responses into revisions and justifications}
\label{sec:appendix_split_algorithm}

Here we describe our heuristic algorithm for extracting the revised sentences and justifications from model output, and present an evaluation of this algorithm’s accuracy.

To identify the revised sentence, we first split model output into sentences using NLTK’s sentence tokenizer (3.9.1).
We take the revision to be the sentence that is most similar to the input sentence stimulus, based on METEOR scores \citep{banerjee2005meteor}.
Because the input sentence may be split into multiple sentences during revision, we also consider sequences of up to 3 contiguous sentences as possible revisions.
We exclude sentences that are identical to the input sentence, as sometimes model outputs repeat the input sentence before the revised version.
We take the rest of the response following the revised sentence to be the justification.

In some cases, models proposed multiple possible revisions. 
We aimed to select the first proposed revision, by removing any text following the phrase “option 2” before running the algorithm described above.
This kind of response was particularly common for the gemma model.

To evaluate the accuracy of our heuristic algorithm, we randomly sampled $n=50$ responses per model, which were not considered in developing our heuristics. 
The algorithm achieves an average accuracy of $94\%$ in exactly identifying revised sentences and $93\%$ in exactly identifying justifications. 
See accuracy per model in Table \ref{tab:split_algo_acc}.

\begin{table}
    \small
    \centering
    \rowcolors{1,3}{}{gray!10}
    \begin{tabular}{*3c}
        \toprule
        & Revision & Justification \\    
        \midrule
        gemma-2-9b-it               & 86  & 82 \\
        gpt-4o                      & 94  & 94 \\
        llama-3.1-8B-Instr.       & 96  & 94 \\
        Mistral-Nemo-Instr.  & 100 & 100 \\
        \midrule
        Overall                     & 94 & 93 \\
        \bottomrule
    \end{tabular}
    \caption{Accuracy of our heuristic algorithm. (percentage correctly identified)}
    \label{tab:split_algo_acc}
\end{table}
\section{Alternative wording revisions}
\label{sec:app_alternative_wording}

\begin{table}
    \centering
    \small
    \rowcolors{1,3}{}{gray!10}
    \begin{tabular}{cc}
        \toprule
        starting variant & percentage gender-neutral \\    
        \midrule
        neutral         & 95   \\
        feminine        & 96   \\
        masculine       & 96   \\
        \midrule
        Overall         & 95.7  \\
        \bottomrule
    \end{tabular}
    \caption{Rates of gender-neutral alternative wordings, by starting variant}
    \label{tab:alt_wording_gend_neutral}
\end{table}

\begin{table*}
    \centering
    \small
    \rowcolors{1,3}{}{gray!10}
    \begin{tabular}{ccccccc}
        \toprule
        starting variant & alt. noun phrase & removed & mention of profession & verb phrase & other & N/A \\    
        \midrule
        neutral          &    44 &     9 &    10 &    7 &    24 &     6 \\
        feminine         &    59 &    10 &    10 &    6 &    12 &     3 \\
        masculine        &    58 &    12 &     9 &    6 &    11 &     4 \\
        \midrule
        Overall          &    53.7 &   10.3  &    9.7 &    6.3 &    15.7 &  4.3    \\
        \bottomrule
    \end{tabular}
    \caption{Sub-categories of gender-neutral alternative wordings, by starting variant. (Cells present percentages of alternative wordings that fall into each sub-category.)}
    \label{tab:alt_wording_types}
\end{table*}

In Sections \ref{sec:revisions} and \ref{sec:justifications}, we split model revisions into four types, based on what role noun variants were revised to: neutral, feminine, masculine, and ``alternative wording.''
Because alternative wordings make up such a large share of revisions, we need to understand their make-up.
Here we assess:

\begin{enumerate}
    \item Are alternative wordings typically gender-neutral?
    \item What are common sub-categories of alternative wordings?
\end{enumerate}

Additionally, in Sec \ref{sec:justifications}, we compare justifications across different starting role noun variants (i.e., whether the role noun in the input sentence was neutral, masculine, or feminine), when revising to alternative wordings (H1b).
Because of this, it is also important to understand whether the qualities of alternative wordings vary across starting variants, which would inform our interpretation of results.
So, for each of the questions above, we also assess whether we observe differences across starting variants (in rates of gender-neutral alternative wordings for 1, and in frequency of sub-categories of alternative wordings for 2).

\subsection{Rates of gender-neutral alternative wordings}

To assess rates of gender-neutral alternative wordings, we randomly sampled 75 responses per model, split evenly across starting variants, from the subset of responses considered in the justifications analysis (300 responses total).
We manually annotated these responses to assess whether the revision was gender-neutral (i.e., did not introduce lexically gendered words).
The vast majority of revisions were gender-neutral ($95.7\%$), with similar rates across starting variants, as shown in Table \ref{tab:alt_wording_gend_neutral}.

Most of the gendered alternative wordings involved using a word that was morphologically related to the role noun (e.g., revising \textit{craftsman} to instead talk about \textit{craftsmanship}, or revising \textit{gentleperson} to talk about someone’s \textit{gentlemanly nature}).
Gender-neutral alternative wordings were quite varied, for example, 
replacing \textit{ambulancewoman} with \textit{paramedic}; 
replacing \textit{fanboy} with \textit{enthusiast}; 
replacing \textit{newswoman} with \textit{freelance journalist}; 
replacing \textit{businessperson} with talking about \textit{leading businesses}; and 
replacing \textit{spokesperson} with talking about \textit{advocating} for something.
We go into greater depth about the make-up of alternative wordings in the next subsection.

The high rates of gender-neutral alternative wordings motivate treating this category as gender-neutral.
Additionally, similar rates of gender-neutral alternative wordings across starting variants supports comparing their associated justifications to assess H1b.

\subsection{Make-up of alternative wordings}

In addition to understanding the rate of gender-neutral alternative wordings, we also wanted to get a general sense of their make-up.
We used an inductive coding approach to develop a categorization scheme for alternative wordings, and identified the following generalizable categories:

\begin{enumerate}
    \item \textbf{Alternative Noun Phrase}: The role noun is replaced by a noun phrase not present in our role noun set (e.g., \textit{outdoorsperson} → \textit{outdoor enthusiast}).
    \item \textbf{Removed}: The role noun is entirely omitted without replacement.
    \item \textbf{Mentions of Profession}: The role noun is substituted with a description explicitly referencing the field or profession (e.g., \textit{firefighter} → \textit{work in firefighting} or  \textit{businessperson} → \textit{career in business}).
    \item \textbf{Verb Phrase}: The intended meaning of the original role noun is conveyed through a verb phrase describing associated actions or responsibilities, rather than naming the role directly (e.g., revising to replace \textit{outdoorsperson} with a phrase talking about \textit{exploring the great outdoors}).
    \item \textbf{Other}: Revisions that do not clearly fit into any of the categories above. Some examples include metaphorical uses of the role noun (e.g., \textit{work like a madman} → \textit{work tirelessly}) and substitutions with placeholders (e.g., \textit{postwoman} → \textit{[insert his profession here]}).
    \item \textbf{N/A}: Cases where the split algorithm from Appendix \ref{sec:appendix_split_algorithm} did not correctly identify the revised sentence.
\end{enumerate}

\noindent
Two authors used this scheme to annotate the same sample from the previous subsection, and then discussed to resolve any disagreements.

The breakdown of alternative wording types by starting variant is shown in Table \ref{tab:alt_wording_types}.
For all starting variants, the most frequent alternative wording sub-category is alternative noun phrases.
One difference across variants is that alternative noun phrases appear slightly more frequent for gendered starting variants, compared to neutral ones.
However, in general, the frequency of the categories across starting variants has a similar distribution, motivating comparing their associated justifications in H1b.

\section{Word sets for justifications analysis}
\label{sec:app-seed-sets}

To study the presence of different themes in justifications, we required word sets corresponding to each theme.
We started with manually curated seed sets.
Half the words in each seed set were synonyms of the theme label word, and half were antonyms. For example, for the theme \texttt{inclusive}, the seed set was \textit{\{inculsive, exclusionary\}}. Seed words for each theme are italicized in Table \ref{tab:seed_sets}.

Then, we used contextual word embeddings from BERT \citep{devlin2018bert} to build expanded sets of 10 words per theme, based on these seed sets.
We started by identifying sentences in the justifications that mention role nouns.
We identified adjectives that occur in these sentences, using spaCy's part of speech tagger. 
We forced the inclusion of some frequent hyphenated adjectives that were split into multiple tokens by spaCy (\textit{gender-neutral}, \textit{gender-specific}, and \textit{non-binary}), resulting in $N=1,039$ total adjectives.
We then generated contextual embeddings using BERT (specifically \texttt{bert-base-uncased}) for each adjective token.
Since we were specifically interested in representing value-relevant properties of adjectives, rather than information about job roles, we replaced role noun variants with \texttt{[MASK]} tokens, to limit the influence of specific occupations on these representations.
For adjectives that corresponded to multiple wordpiece tokens, we averaged the wordpiece contextual embeddings.

We then created word embeddings per adjective by averaging the contextual embeddings of all of that adjective's occurrences.
Next, we generated theme embeddings by averaging the embeddings of the words in each seed set (e.g., averaging the embeddings of \textit{inclusive} and \textit{exclusionary} for the \texttt{inclusive} theme).
Then, we combined seed sets with the 10 nearest neighbor adjectives for each theme embedding to get the full word sets per theme. 

\section{Descriptive statistics about justification themes}
\label{app:justification_tables}

\begin{table}
    \centering
    \small
    \begin{tabular}{lrrr}
    \toprule
        & \multicolumn{3}{@{}c}{starting role noun gender}\\
    theme & neutral & feminine & masculine \\
    \midrule
    inclusive & 6 & 29 & 29 \\
    modern & 5 & 10 & 12 \\
    professional & 12 & 9 & 11 \\
    standard & 11 & 6 & 6 \\
    natural & 8 & 4 & 5 \\
    \bottomrule
    \end{tabular}
\caption{Themes in justifications by starting variant. Cells indicate the percentage of justifications where a theme was mentioned. Based on $13,609$  responses, where role nouns were revised to alternative wordings.}
\label{tab:just_props_variants}
\end{table}

\begin{table*}
    \centering
    \small
    \begin{tabular}{l|rrr|rrr|rrr}
    \toprule
        & \multicolumn{3}{@{}c}{gender declaration} 
        & \multicolumn{3}{c@{}}{pronoun declaration}
        & \multicolumn{3}{c@{}}{pronoun usage}\\

    theme & nonbinary & woman & man & they/them & she/her & he/him & their & her & his \\
    \midrule
    inclusive & 58 & 48 & 38 & 56 & 65 & 54 & 41 & 44 & 43 \\
    modern & 6 & 18 & 17 & 8 & 22 & 17 & 21 & 25 & 22 \\
    professional & 4 & 20 & 17 & 4 & 13 & 9 & 23 & 19 & 20 \\
    \bottomrule
    \end{tabular}
\caption{Themes in justifications for revisions to neutral. Cells indicate the percentage of justifications where a theme was mentioned. Based on $6,964$ model responses, where role nouns were revised from gendered to neutral.}
\label{tab:just_props_genders}
\end{table*}

Table \ref{tab:just_props_variants} shows the breakdown of themes across starting role noun variants, and Table \ref{tab:just_props_genders} shows the breakdown of themes across preambles.

\end{document}